\documentclass{article} %
\usepackage{iclr2025_conference,times}
\usepackage{amsthm}
\usepackage{algorithm}
\usepackage[noend]{algpseudocode}
\usepackage{algpseudocode}
\usepackage{courier}
\usepackage{graphicx}
\usepackage{mathtools}
\usepackage{setspace}
\usepackage{tabularx}
\usepackage{subfigure}
\usepackage{sistyle}
\SIthousandsep{,}
\usepackage{svg}
\svgpath{{figures/}{real_figures/}} %
\usepackage{wrapfig}

\usepackage{amsmath,amsfonts,bm}
\usepackage{xspace}

\def\eqref#1{equation~\ref{#1}}

\def\floor#1{\lfloor #1 \rfloor}
\def\1{\bm{1}}

\DeclareMathAlphabet{\mathsfit}{\encodingdefault}{\sfdefault}{m}{sl}
\SetMathAlphabet{\mathsfit}{bold}{\encodingdefault}{\sfdefault}{bx}{n}

\newcommand{\Var}{\mathrm{Var}}

\newcommand{\Cov}{\mathrm{Cov}}

\newcommand{\iid}{i.i.d.\@\xspace}

\usepackage{hyperref}
\usepackage[noabbrev]{cleveref}
\usepackage{url}
\newcommand{\customstrut}{\rule[0pt]{0pt}{0pt}}

\title{Infinite-Resolution Integral Noise Warping \\for Diffusion Models}

\author{Yitong Deng$^{1,2}$, Winnie Lin$^{1}$, Lingxiao Li$^{1}$, Dmitriy Smirnov$^{1}$, Ryan Burgert$^{3,4}$, Ning Yu$^{3}$,\\
\textbf{Vincent Dedun$^{1}$, Mohammad H. Taghavi$^{1}$}\\
$^{1}$Netflix, $^{2}$Stanford University, $^{3}$Netflix Eyeline Studios, $^{4}$Stony Brook University\\
\texttt{yitongd@stanford.edu} \\
\texttt{\{winniel, lingxiaol, dimas, vdedun, mtaghavi\}@netflix.com} \\
\texttt{rburgert@cs.stonybrook.edu}\\
\texttt{ning.yu@scanlinevfx.com}
}

\iclrfinalcopy %
\theoremstyle{definition}

\newtheorem{theorem}{Theorem}

\algnewcommand\algorithmicforeach{\textbf{for each}}
\algdef{S}[FOR]{ForEach}[1]{\algorithmicforeach\ #1\ \algorithmicdo}
\algnewcommand\algorithmicparallelforeach{\textbf{parallel for each}}
\algdef{S}[FOR]{ParallelForEach}[1]{\algorithmicparallelforeach\ #1\ \algorithmicdo}

\begin{document}

\newif\ifcomments
\commentstrue

\ifcomments
\newcommand{\yitong}[1]{\textcolor{red}{[yitong: #1]}}
\newcommand{\lingxiao}[1]{\textcolor{blue}{[lingxiao: #1]}}
\newcommand{\dima}[1]{\textcolor{green}{[dima: #1]}}
\newcommand{\winnie}[1]{\textcolor{orange}{[winnie: #1]}}
\else
\newcommand{\yitong}[1]{}
\newcommand{\lingxiao}[1]{}
\newcommand{\dima}[1]{}
\newcommand{\winnie}[1]{}
\fi

\maketitle
\vspace{-10pt}
\begin{abstract}
Adapting pretrained image-based diffusion models to generate temporally consistent videos has become an impactful generative modeling research direction.
Training-free noise-space manipulation has proven to be an effective technique, where the challenge is to preserve the Gaussian white noise distribution while adding in temporal consistency.
Recently, \citet{chang2024warped} formulated this problem using an integral noise representation with distribution-preserving guarantees, and proposed an upsampling-based algorithm to compute it. However, while their mathematical formulation is advantageous, the algorithm incurs a high computational cost. 
Through analyzing the limiting-case behavior of their algorithm as the upsampling resolution goes to infinity, we develop an alternative algorithm that, by gathering increments of multiple Brownian bridges, achieves their infinite-resolution accuracy while simultaneously reducing the computational cost by orders of magnitude.
We prove and experimentally validate our theoretical claims, and demonstrate our method's effectiveness in real-world applications. We further show that our method readily extends to the 3-dimensional space. 
\end{abstract}

\section{Introduction}

The success of diffusion models in image generation and editing \citep{rombach2022high,nichol2021glide,ho2020denoising,zhang2023adding} has spurred significant interest in lifting these capacities to the video domain \citep{singer2022make,durrett2019probability,gupta2023photorealistic,blattmann2023videoldm,ho2022imagen,guo2023animatediff}.
While training video diffusion models directly on spatiotemporal data is a natural idea, practical concerns such as limited availability of large-scale video data and high computational cost have motivated investigations into training-free alternatives. One such approach is to use pre-trained image models to directly generate video frames, and utilize techniques such as cross-frame attention, feature injection and hierarchical sampling to promote temporal consistency across frames \citep{ceylan2023pix2video, zhang2023controlvideo,khachatryan2023text2video,cong2023flatten}. %

Among these techniques, the controlled initialization of noise has been consistently shown to be an important one \citep{ceylan2023pix2video, khachatryan2023text2video}. However, most existing approaches for noise manipulation either compromise the noise Gaussianity (and subsequently introduce a domain gap at inference time), or are restricted to simple manipulations such as filtering and blending which are insufficient for capturing complex temporal correlations.
Recently, \citet{chang2024warped} 
proposed a method that both preserves Gaussian white noise distribution and well captures temporal correlations via \emph{integral noise warping}: each warped noise pixel integrates a continuous noise field over a polygonal deformed pixel region, which is computed by summing subpixels of an upsampled noise image. However, their method's theoretical soundness and effectiveness are followed by its high-end computational cost in both memory and time, which not only incurs a significant overhead at inference time but also limits its useability in novel applications \citep{kwak2024geometry}.

In this paper, we introduce a new noise-warping algorithm that dramatically cuts down the cost of \citet{chang2024warped} while fully retaining its virtues. Our key insight for achieving this lies in that, when adopting an Eulerian perspective (as opposed to the original Lagrangian one), the limiting-case algorithm of \citet{chang2024warped} for computing a warped noise pixel reduces to summing over increments from multiple Brownian bridges \citep[Section 8.4]{durrett2019probability}.
In place of the costly upsampling procedure, sampling the increments of a Brownian bridge can be done efficiently in an autoregressive manner (\ref{eq:conditional_scalar_dist}).
We build upon this to devise the \emph{infinite-resolution integral noise warping} algorithm (\ref{alg:main_algo})
which directly resolves noise transport in the continuous space, when given an oracle that returns the overlapping area between a pixel square and a deformed pixel region (\Cref{subsec:partition}).

We propose two concrete ways to compute this oracle, leading to a \textit{grid-based} and a \textit{particle-based} variant of our method. Similar to \citet{chang2024warped}, the \textit{grid-based} variant (Algorithm~\ref{alg:mesh_based}) computes the area by explicitly constructing per-pixel deformed polygons, and is exactly equivalent to the existing approach \citep{chang2024warped} with an infinite upsampling resolution, while running $8.0\times$ to $19.7\times$ faster and using $9.22\times$ less memory\footnote{Since the official code of \cite{chang2024warped} is not available, performance is compared using our reimplementation in Taichi \citep{hu2019taichi}, which we find to be faster than as reported in the original paper.}. Inspired by hybrid Eulerian-Lagrangian fluid simulation \citep{brackbill1988flip}, our novel \textit{particle-based} variant (Algorithm~\ref{alg:mesh_free}) computes area in a fuzzy manner, which not only offers a \textit{further} $5.21\times$ speed-up \textit{over our grid-based variant}, but is also agnostic to non-injective maps. In real-world scenarios, the particle-based variant shows no compromise in generation quality compared to the grid-based one (see video results), while offering superior robustness, efficiency, simplicity, and extensibility to higher dimensions.

In summary, we propose a new noise-warping method to facilitate video generation by lifting image diffusion models. Through analyzing the limiting case of the current state-of-the-art method \citep{chang2024warped} with an infinite upsampling resolution, we derive its continuous-space analogy, which fully retains its distribution-preserving and temporally-coherent properties, while achieving orders-of-magnitude speed-up, warping $1024 \times 1024$ noise images in $\sim 0.045$s (grid variant) and $\sim 0.0086$s (particle variant) using a laptop with a Nvidia RTX 3070 Ti GPU.

\begin{figure}[!t]
    \vspace{-15pt}
    \begin{subfigure}
        \centering
        \includegraphics[width=0.98\linewidth]{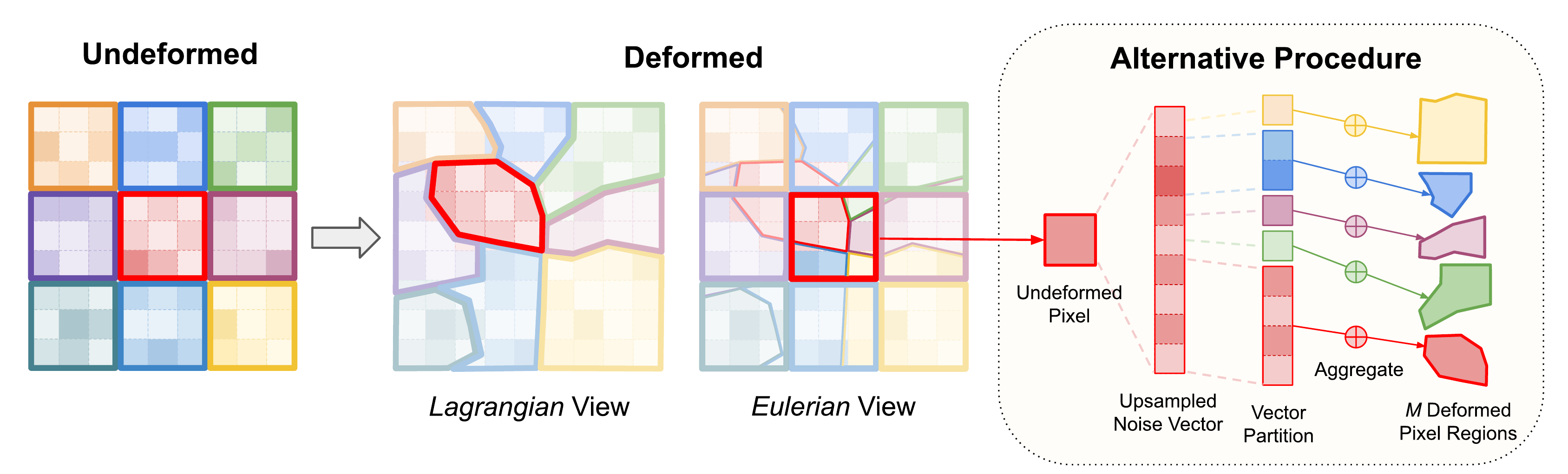}
    \end{subfigure}
    \vspace{-0.08in}
    \caption{When the image grid deforms, the Lagrangian view tracks a deformed pixel region, while the Eulerian view tracks the undeformed pixel square as it gets partitioned into multiple regions. On the right, we leverage the exchangeability of upsampled subpixels to convert the Lagrangian gathering procedure into scattering noise subpixels to overlapped deformed pixel regions.
    }
    \label{fig: eulerian_viewpoint}
    \vspace{-5pt}
\end{figure}

\section{Methodology}
In this section, we introduce our method as follows: 
\begin{itemize}
    \item %
    We present an equivalent Eulerian interpretation (\Cref{fig: eulerian_viewpoint}) for the method by \cite{chang2024warped}, which was developed from a Lagrangian viewpoint.
    \item  %
    We show that the limiting algorithm of the Eulerian formulation as upsampling level goes to infinity is equivalent to sampling increments of Brownian bridges.
    \item %
    We present our main algorithm (\ref{alg:main_algo})
    which, given a partition record that returns the overlapping area between a pixel square and a deformed pixel region, samples increments of Brownian bridges and scatters the increments to form the warped noise image.
    \item %
    We propose two concrete algorithms for computing the overlap areas. The \textit{grid-based} Algorithm~\ref{alg:mesh_based} extends \citet{chang2024warped} to infinite resolution without the overhead of upsampling. 
    The \textit{particle-based} Algorithm~\ref{alg:mesh_free} departs from grid-based discretization and uses particles instead, resulting in a simpler algorithm that is robust to degenerate maps.
\end{itemize}

\subsection{Noise Warping: an Alternative Eulerian Perspective}

Given a $D \times D$ prior
noise image $I_W \in \mathbb{R}^{D\times D}$\footnote{
Here we assume that the noise image is square and has a single channel only to simplify notation. In practice, the noise image can have arbitrary aspect ratio and number of independent channels.} and a deformation map $\psi : [0, 1]^2\to [0,1]^2$, the noise-warping algorithm \citep{chang2024warped} computes the warped noise image $\widetilde{I}_W \in \mathbb{R}^{D\times D}$ with upsampling level $N \in \mathbb{Z}_{\ge 1}$ as follows:
\begin{enumerate}
    \item For $i,j=1,\ldots,D$, upsample noise pixel $[I_{W}]_{i,j}$ to an $N \times N$ subimage $[\widehat{I}_W]_{i,j} \in \mathbb{R}^{N\times N}$: %
    \begin{align}
        [\widehat{I}_{W}]_{i,j} %
        = \frac{[I_{W}]_{i,j}}{N^2} + \frac{1}{N}\!\left(\!Z\!-\!\frac{S}{N^2}\!\right)\!,\text{ with } Z \sim \mathcal{N}(\mathbf{0}, \mathbf{I}) \text{ and }S = 
\sum\nolimits_{k=1}^{N^2}Z_k.
        \label{eq: cond_distribution_image}
    \end{align}
    The subimage for each pixel assembles into an $ND \times ND$ upsampled noise image $\widehat{I}_W$.
    \item For $i,j=1,\ldots,D$, the pixel square $A_{i,j} \coloneq [\frac{i-1}{D},\frac{i}{D}]\times [\frac{j-1}{D}, \frac{j}{D}]$ is warped to a deformed pixel region $\widetilde{A}_{i,j} \coloneq \psi(A_{i,j})$, and the warped noise pixel $[\widetilde{I}_W]_{i,j}$ is set to be the sum of all subpixels in $\widehat{I}_W$ covered by $\widetilde{A}_{i,j}$ divided by $\sqrt{|\widetilde A_{i,j}|}$, where $|A|$ denotes the Lebesgue measure of a Borel set $A \subset \mathbb{R}^2$. 
\end{enumerate}

We describe an alternative but equivalent procedure by making the following two observations, which are illustrated in Figure~\ref{fig: eulerian_viewpoint}.

\textbf{Gathering Noise $\rightarrow$ Scattering Noise.}
While the original procedure computes the warped noise image by \emph{gathering} the upsampled noise subpixels in each deformed pixel region $\widetilde{A}_{i,j}$ in a \emph{Lagrangian} fashion, we can instead use an alternative procedure by \emph{scattering} the upsampled noise subpixels in each pixel square $A_{i,j}$ to overlapping deformed pixel regions.
This new \emph{Eulerian} procedure does not change the output, but it yields new insights in conjunction with our second observation.

\textbf{Scattering Noise $\rightarrow$ Counting Overlapping Subpixels.}
Observe that the $N\times N$ subpixels in $[\widehat I_W]_{i,j}$, for every $i,j$, are correlated only through their sum $S$  when conditioning on $[I_W]_{i,j}$ (\ref{eq: cond_distribution_image}), so they are exchangeable.
Hence, when scattering these upsampled noise subpixels to deformed pixel regions, the order of scattering does not matter, and we only need to count \emph{the number of subpixels} covered by each deformed pixel region.

\textbf{Alternative Eulerian Procedure.} 
Putting both observations together, we now describe an alternative procedure to \citet{chang2024warped} with unaltered output:
\begin{enumerate}
\item For each noise image pixel $[I_{W}]_{i,j}$, draw an upsampled subimage, now represented as a 1D vector $X \in \mathbb{R}^{N^2}$ using (\ref{eq: cond_distribution_image}). Then, compute a prefix sum $H_{i,j}$ via $[H_{i,j}]_k \coloneq \sum_{q=1}^{k} X_q$ for $k=1,\ldots,N^2$.
\item Warp each pixel square and compute deformed pixel regions $\widetilde A_{i,j}$ as before.
\item For each $A_{i,j}$, identify all $M$ deformed pixel regions $\{\widetilde A_{\ell_k,m_k}\}_{k=1,\ldots, M}$ that overlap with $A_{i,j}$. Form $L \in \mathbb{Z}_{\ge 0}^{M}$ where $L_k$ represents the number of upsampled subpixels covered by the $k^\text{th}$ overlap. Then, compute a prefix sum $[C_{i,j}]_k \coloneq \sum_{q=1}^{k} L_q$. For $k= 1, \ldots, M$, accrue $[H_{i,j}]_{[C_{i,j}]_k} - [H_{i,j}]_{[C_{i,j}]_{k-1}}$ to $[\widetilde I_{W}]_{\ell_k, m_k}$, the $k^\text{th}$ overlapped warped noise pixel.
\item  Divide each warped noise pixel $[\widetilde I_W]_{i,j}$ by $\sqrt{|\widetilde A_{i,j}|}$.
\end{enumerate}

\textbf{Discussion.} 
Compared to the original procedure by \citet{chang2024warped}, this alternative but equivalent algorithm highlights how the upsampled subpixels of $[I_W]_{i,j}$ are scattered to form the warped noise pixels.
In particular, each warped noise pixel receives the \emph{sum of a continuous segment} in $H_{i,j}$.
Since $H_{i,j}$ is a summation of weakly correlated and exchangeable subpixels, once conditioned on $[I_W]_{i,j}$, \emph{can we avoid explicitly instantiating every single subpixel}, but instead model the \emph{sum} of these weakly correlated subpixels?

The key insight of this paper is that when the upsampling resolution $N \to \infty$, the scaling limit of the prefix sum $H_{i,j}$ (with proper interpolation and time scaling to a continuous function) is precisely the Brownian bridge \citep[Section 8.4]{durrett2019probability} conditioned on $[I_W]_{i,j}$. 
Once this connection is established, it is easy to progressively sample increments of the Brownian bridge, resulting in a clean and efficient noise-warping algorithm that bypasses the need for upsampling in \citet{chang2024warped}.

\begin{figure}[!t]
    \vspace{-25pt}
    \begin{subfigure}
        \centering
        \includegraphics[width=0.98\linewidth]{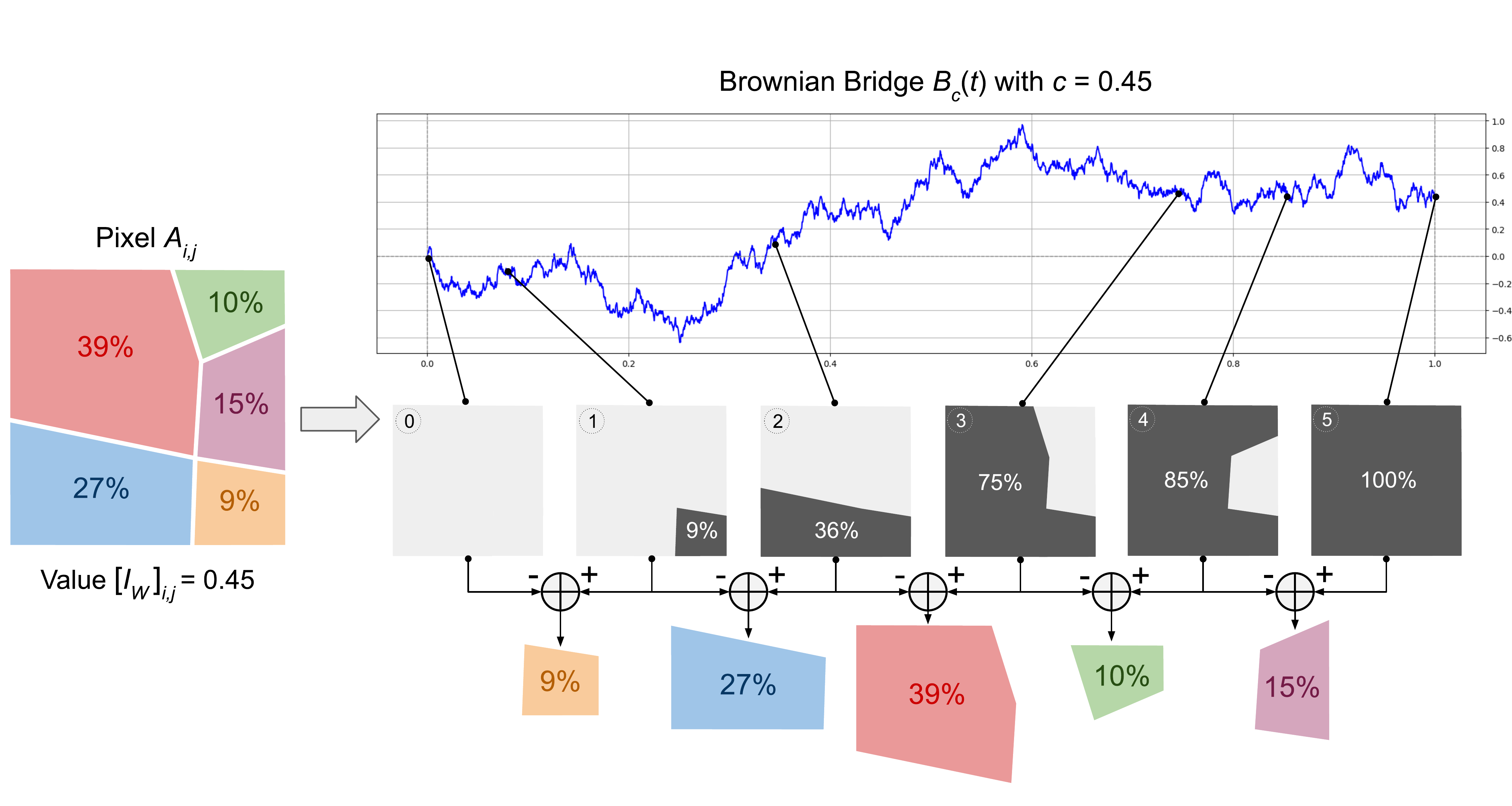}
    \end{subfigure}
    \vspace{-0.2in}
    \caption{Connection between Eulerian noise-warping and increments of a Brownian bridge for a fixed prior noise pixel $[I_W]_{i,j}$. 
    The overlapping area of each colored warped region becomes the time increment for the Brownian bridge. Hence, sampling the Brownian bridge at these times and taking consecutive differences yields integral noise that is  scattered to form each warped noise pixel.}
    \label{fig: brownian_bridge_sampling}
    \vspace{-4pt}
\end{figure}

\subsection{Infinite-Resolution Noise Scattering}
In this section, we first derive a scaling limit result to Brownian bridges. We then illustrate that the limiting version of the Eulerian procedure from the previous section matches precisely this scaling limit result.
Lastly, we describe an autoregressive way to sample increments of a Brownian bridge that is linear in runtime in terms of the number of increments.

\begin{theorem}[Scaling limit to Brownian bridge]
\label{theo: convergence}
Let $\{Z_n\}$ %
be a sequence of \iid %
random variables with finite variance that are normalized such that $\mathbb{E}[Z_n] = 0$ and $\Var(Z_n) = 1$.
For $c \in \mathbb{R}$, define
\begin{align*}
    S_n \coloneq \sum_{i=1}^n Z_i, \qquad X_{i,n} \coloneq \frac{c}{n} + \frac{1}{\sqrt{n}}\left(Z_i - \frac{S_n}{n}\right).
\end{align*}
Consider the sequence of random continuous functions $\{H_n(t)\} %
\subset C[0, 1]$ defined as
\begin{align*}
H_n(t) \coloneq \sum_{i=1}^{\floor{nt}} X_{i,n} + (nt-\floor{nt})X_{\floor{nt}+1, n}.
\end{align*}
Then the sequence $\{H_n\}$ %
converges in distribution under the sup-norm metric on $C[0,1]$ to 
$B_c(t) \coloneq W(t) - tW(1) + tc$,
the Brownian bridge ending at $c$, where $W(t)$ is standard Brownian motion.
Moreover, in distribution, we have
$B_c(t) \stackrel{d}{=} (W(t) \mid W(1) = c), \label{eqn: BB_cond_thm}$
where  $(W(t) \mid W(1) = c)$ is %
the disintegrated measure \citep{pachl1978disintegration} of $W(t)$ on $W(1)=c$.
\end{theorem}
We prove Theorem~\ref{theo: convergence} in Appendix~\ref{append: proof1}.
To connect the Eulerian procedure with the setup in Theorem~\ref{theo: convergence}, let us fix a pixel $[I_W]_{i,j}$, and let $B \coloneq B_{[I_W]_{i,j}}$, $H \coloneq H_{i,j}$, $C \coloneq C_{i,j}$ to simplify the notation. By setting $n = N^2$ and $c = [I_W]_{i,j}$, the sequence $\{X_{k,n}\}$ from the theorem has exactly the same law as the upsampled subpixels in $[\widehat I_W]_{i,j}$.
Moreover, $H_{nt} = H_n(t)$ when $nt \in \mathbb{Z}_{\ge 1}$.
By taking $N \to \infty$, implying $n \to \infty$, for any $t_1,\ldots, t_M \in [0, 1]$, we have the convergence in distribution of
    $\left(H_{\floor{n {t_1}}}, \ldots, H_{\floor{n {t_{M}}}}\right) \stackrel{d}{\to} \left(B(t_1), \ldots, B(t_M)\right)$.
Recall in the Eulerian procedure, we only need to access the prefix sum $H$ at indices $\{C_k\}_{k=1}^{M}$, where $C_k$ counts the number of upsampled subpixels covered by the first $k$ overlaps.
This suggests that if we choose 
\begin{align*}
t_k = \lim_{N\to\infty} \frac{C_k}{N^2} = \sum_{k'= 1}^k\left|A_{i,j} \cap \widetilde A_{\ell_{k'},m_{k'}}\right|,
\end{align*} 
and use $B(t_k)$ in place of $H_k$,
then we just need to sample from %
$B$ at times $t_1,\ldots,t_{M}$ --- precisely the limiting algorithm of the Eulerian procedure.
We illustrate this connection in \Cref{fig: brownian_bridge_sampling}.

\textbf{Autoregressive Sampling of Brownian Bridges.}
Since a Brownian bridge is a Markov process \citep[Exercise 5.11]{oksendal2013stochastic}, we can sample the vector $(B_c(t_1),\ldots, B_c(t_M))$ in an autoregressive fashion, each time sampling $B_c(t_{k+1})$ conditioned on $B_c(t_{k})$: %
\begin{align}
\left(B_c(t_{k+1}) \mid B_c(t_k) = q\right) \stackrel{d}{=} \operatorname{\mathcal N} \left(
\frac{1-t_{k+1}}{1-t_{k}} q + \frac{t_{k+1}-t_k}{1-t_k} c,
\frac{(t_{k+1}-t_k)(1-t_{k+1})}{1-t_k}
\right).
    \label{eq:conditional_scalar_dist} 
\end{align}
Once the Brownian bridge at times $t_k$ is sampled, we just need to accrue the increments $B_c(t_{k}) - B_c(t_{k-1})$  to $[\widetilde I_W]_{\ell_k, m_k}$, the $k^{\text{th}}$ overlapped warped noise pixel.
This allows us to present \Cref{alg:main_algo}.
Compared to the discrete procedures described earlier, we no longer need upsampling. In addition, we exploited the autoregressive nature of Brownian bridges to bring down the time complexity to linear in the number of overlapping warped pixel regions.

\begin{algorithm}
\caption{Infinite-Resolution Integral Noise Warp}\label{alg:main_algo}
\begin{algorithmic}
\Require prior noise image $I_W \in \mathbb{R}^{D\times D}$, deformation map $\psi: [0,1]\to[0,1]$
\Ensure warped noise image $\widetilde I_W\in \mathbb{R}^{D\times D}$
\State Build a partition record $\mathcal{P}$ from $\psi$ (\Cref{subsec:partition})
\State Initialize $\mathcal{A}_{i,j} \gets 0$ for all $i,j = 1,\ldots,D$ \Comment{$\mathcal{A}_{i,j}$ will eventually be the area of $\widetilde A_{i,j}$}
\ParallelForEach {$u,v = 1,\ldots, D$}
    \State $t, q, M \gets 0, 0, |\mathcal{P}_{u,v}|$
    \For {$k = 1,\ldots, M$}
    \State $(a, i, j) \gets [\mathcal{P}_{u,v}]_k$ \Comment{$a$ is the overlapping area between $A_{i,j}$ and $\widetilde A_{u,v}$}
    \State Sample $q' \sim (B_c(t+a) | B_c({t}) = q)$ by (\ref{eq:conditional_scalar_dist}) with $c = [I_W]_{u,v}$
    \State $[\widetilde{I}_W]_{i,j} \gets [\widetilde{I}_W]_{i,j} + (q'-q)$
    \State $\mathcal{A}_{i,j} \gets \mathcal{A}_{i,j} + a$
    \State $q, t \gets q', t + a$
    \EndFor
\EndFor
\State Normalize $[\widetilde{I}_W]_{i,j} \gets {{\mathcal{A}_{i,j}}}^{-\frac{1}{2}}[\widetilde{I}_W]_{i,j}$ for all $i,j = 1,\ldots,D$
\State \Return $\widetilde{I}_W$
\end{algorithmic}
\end{algorithm}

\begin{figure}[!t]
    \vspace{-20pt}
    \centering
    \includegraphics[width=0.98\linewidth]{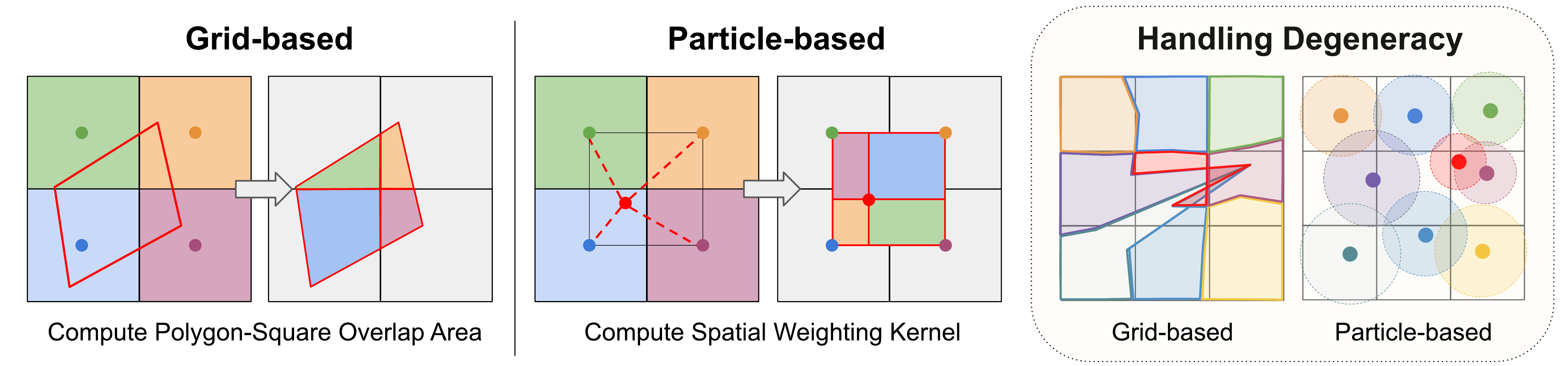}
    \vspace{-0.05in}
    \caption{
    The grid-based variant (left) computes the overlapping areas by explicitly constructing the polygon for the deformed pixel region.
    The particle-based variant (middle) approximates these areas with a weighting kernel. With degenerate maps (right), the fixed topology of the grid-based variant can lead to problems, while the connectivity-free, particle-based variant remains stable.}
    \label{fig:flesh}
    \vspace{-2pt}
\end{figure}

\textbf{Preservation of Gaussian White Noise.}
A central desideratum of noise warping is that the resulting warped noise image $\widetilde I_W$ needs to %
have pixels that are \iid standard Gaussians when the prior noise image $I_W$ is Gaussian white noise.
This ensures that the warped noise is in-distribution for a pre-trained diffusion model.
Our algorithm automatically guarantees this preservation of Gaussianity, as long as the warping function $\psi$ is injective.
To see this, the injectivity of $\psi$ implies that the warped pixel regions are non-overlapping in the square $[0,1]^2$.
For each $A_{i,j}$, since $[I_W]_{i,j} \stackrel{d}{=} \mathcal{N}(0, 1) \stackrel{d}{=} W(1)$, by the conditional interpretation of Brownian bridges (\ref{eqn: BB_cond_thm}), when marginalizing out $[I_W]_{i,j}$, the Brownian bridge $B_{[I_W]_{i,j}}$ reduces to standard Brownian motion.
Since the increments of the Brownian motion are independent Gaussians, the contribution to a deformed pixel region is simply a zero-mean Gaussian with variance equal to the overlapping area.
Therefore, each deformed pixel region will receive the sum of a number of independent Gaussians whose variances sum to the area of the region.
The scaling by the inverse square root of the area in \Cref{alg:main_algo} thus makes each warped noise pixel an \iid standard Gaussian.

\subsection{Building Partition Records}\label{subsec:partition}
To compute \Cref{alg:main_algo}, we need a way to compute the partition record $\mathcal{P}$, which specifies how each pixel square is partitioned by multiple deformed pixel regions.
In this section, we present one grid-based and one particle-based method for building $\mathcal{P}$. In particular, for each pixel square with indices $(u,v)$, we compute $\mathcal{P}_{u,v}$ as a list of 3-tuples $(a, i, j)$, where $(i,j)$ identifies the overlapped deformed pixel region and $a$ represents the overlapping area. Both variants are illustrated in \Cref{fig:flesh}.

\begin{minipage}[t]{0.46\textwidth}
\vspace{-15pt}
\begin{algorithm}[H]
    \centering
    \caption{Grid-based Partition}\label{alg:mesh_based}
    \footnotesize
\begin{algorithmic}
\Require  Deformation map $\psi$
\Ensure  Partition record $\mathcal{P}$ 
\ParallelForEach { $i,j$}
    \State $A^* \gets \text{DiscretizeSquare}(A_{i,j})$
    \State  $S \gets \psi(A^*)$
    \State  $u^-, u^+, v^-, v^+ \gets \text{AABB}(S)$
    \For { $u \in [u^-, u^+]$}
        \For { $v \in [v^-, v^+]$}
        \State  $a \gets \text{PolygonArea}(\text{Clip}(S, u, v))$
        \State  
            $\mathcal{P}_{u,v} \gets \mathcal{P}_{u,v} + [(a, i, j)]$
        \EndFor
    \EndFor
\EndFor
\State  \Return $\mathcal{P}$
\end{algorithmic}
\end{algorithm}
\vspace{1pt}
\end{minipage}
\hfill
\begin{minipage}[t]{0.53\textwidth}
\vspace{-15pt}
\begin{algorithm}[H]
    \centering
\caption{Particle-based Partition}\label{alg:mesh_free}
    \footnotesize
\begin{algorithmic}
\Require  Deformation map $\psi$

\Ensure  Partition record $\mathcal{P}$ 
\ParallelForEach { $i,j$}
    \State  $(x,y) \gets \psi(\frac{i+0.5}{D}, \frac{j+0.5}{D})$
    \State  $w_{0,0}, w_{0,1}, w_{1,0}, w_{1,1} \gets \text{BilinearWeights}(X)$
    \For { $s, t \in [0, 1]$}
        \State  $x', y' \gets \lfloor x \rfloor + s, \lfloor y \rfloor + t$ 
        \State  $\mathcal{P}_{x', y'} \gets \mathcal{P}_{x', y'} + [(w_{s,t}, i, j)]$
    \EndFor
\EndFor
\vspace{-2.6pt}
\ParallelForEach { $u,v$}
    \State \customstrut Normalize total area of $\mathcal{P}_{u,v}$ to ${D^{-2}}$
\EndFor
\State  \Return $\mathcal{P}$
\end{algorithmic}
\end{algorithm}
\end{minipage}

Our grid-based method (\Cref{alg:mesh_based} and \Cref{fig:flesh}, left) follows \cite{chang2024warped} by modeling each deformed pixel region as an octagon and computes overlapping areas by clipping it aginast undeformed pixel squares. %
Our particle-based method (\Cref{alg:mesh_free} and \Cref{fig:flesh}, middle) borrows from the grid-to-particle techniques in fluid particle-in-cell methods \citep{brackbill1988flip}, where we treat each deformed pixel region as a particle and each undeformed pixel square as a grid cell. Each particle requests area from nearby cells based on distance; upon receiving requests, each cell normalizes the requests to ensure partition-of-unity, and distributes its area to contacting particles.

\textbf{Discussion.} 
Conceptually, our grid and particle-based methods correspond to two different interpretations of $\psi$ when provided as discrete samples (\textit{e.g.}, an optical flow image). The grid-based method implicitly reconstructs the continuous $\psi$ field by linear interpolation, whereas the particle-based method assumes $\psi$ is only known point-wise. The implication is that when $\psi$ is smooth, linear interpolation works well and the grid-based method will yield a higher-quality warp as seen in Figure~\ref{fig: mesh_free_diffeomorphic}. But when $\psi$ is non-smooth, which is commonly the case in real world, linear interpolation can lead to degenerate polygons as illustrated on the right of \Cref{fig:flesh}. The spurious overlaps between the degenerate polygons will lead to spatial correlation in the warped noise image.
Although both \cite{chang2024warped} and our grid-based method implement fail-safes\footnote{In our case, we clamp the input $t_{k+1}$ to $1$ before sampling by \ref{eq:conditional_scalar_dist}. Intuitively, this means that when an undeformed pixel has assigned its entire pixel region to deformed pixels, later noise requests will be neglected.} to avoid noise sharing and maintain spatial independence in practice, they suffer from the intrinsic ambiguity caused by these overlaps.  On the other hand, the particle-based method circumvents such overlaps to begin with.

\begin{wrapfigure}[12]{r}[0pt]{0.2\textwidth}  %
    \vspace{-13pt}
    \centering
    \includegraphics[ width=0.99\linewidth]{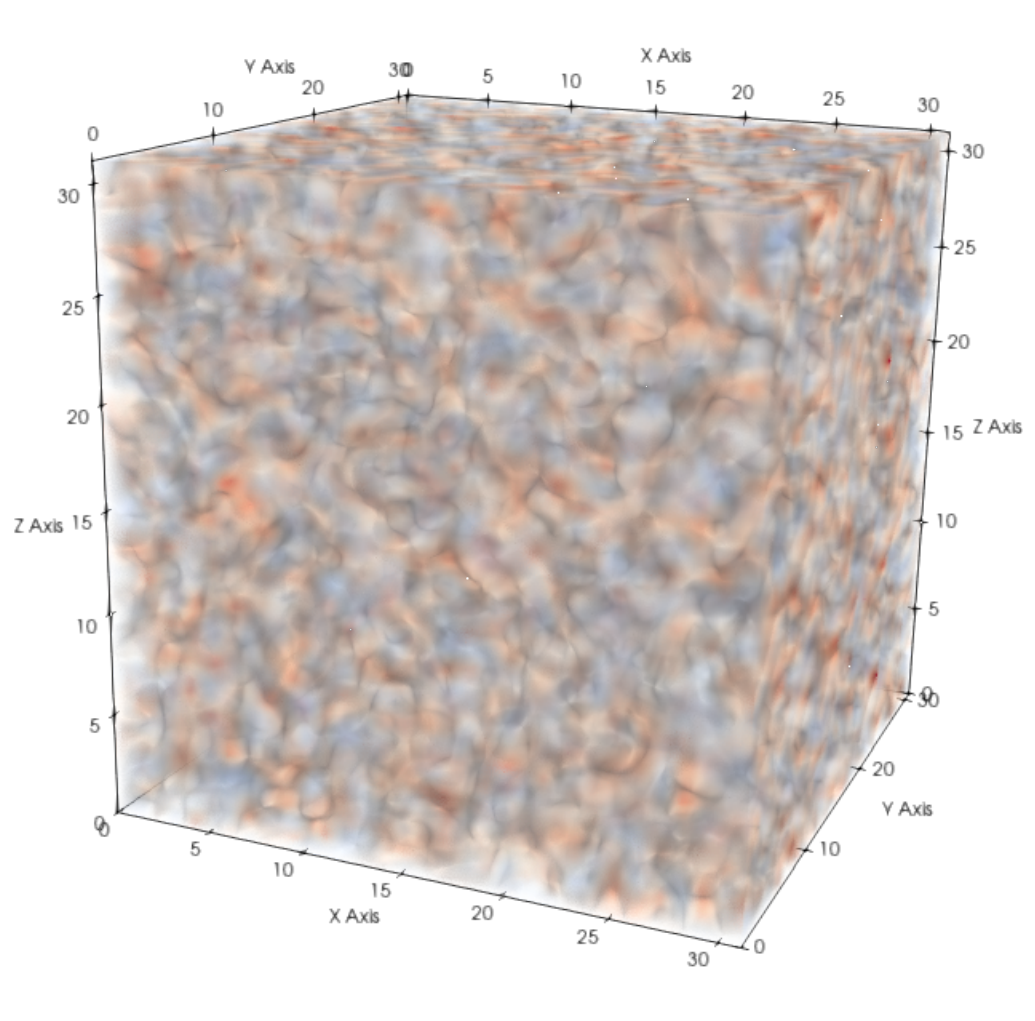}
    \vspace{-15pt}
    \caption{3D noise warped by our particle variant.}
    \label{fig:3d_noise_single}
\end{wrapfigure}

In addition, we highlight the simplicity and parallelizability of the particle-based method, as it boils down the computation of $\mathcal{P}$ to evaluating one bilinear kernel per pixel. Leveraging this fact, we can conveniently and efficiently extend our noise warping algorithm to higher spatial dimensions by replacing the bilinear kernel to its higher-dimensional counterparts, as shown in Figure~\ref{fig:3d_noise_single}.

\section{Results}
In this section, we verify our theoretical claims by showing that our both variants preserve Gaussian white noise distribution, and that \cite{chang2024warped} (HIWYN) converges to our grid-based variant as $N$ increases.
We analyze the behaviors of our grid-based and particle-based variants under diffeomorphic and non-diffeomorphic deformations.
We then apply our method in video generation and benchmark against existing methods \citep{ge2023preserve, chen2023control, chang2024warped}.
Finally, we extend our method to warping volumetric noise and demonstrate a use case in 3D graphics.

\begin{figure}[!t]
    \vspace{-10pt}
    \centering
    \begin{subfigure}
        \centering
        \includegraphics[width=0.99\linewidth]{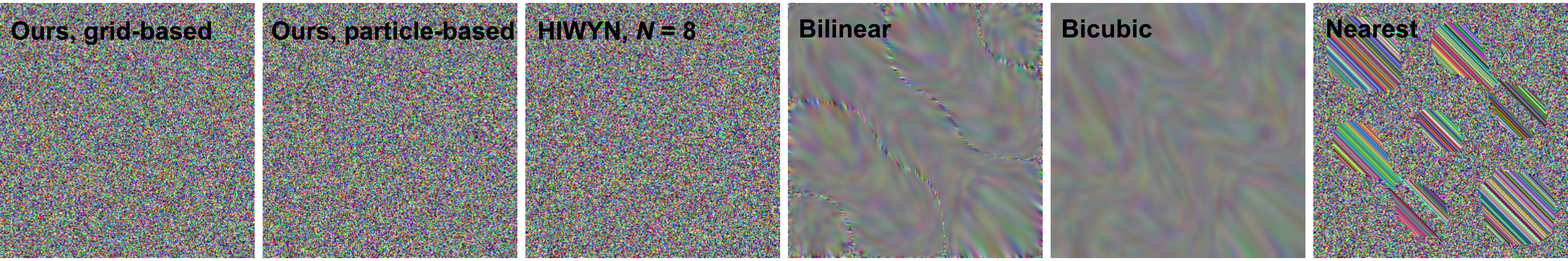}
    \end{subfigure}
\renewcommand{\arraystretch}{1.2}
\scalebox{0.79}{
\begin{tabular}{c|c|c|c|c|c|c}
\hline
\multicolumn{7}{c}{Distribution Preservation Metrics}\\
\hline
    Method & Ours, grid-based & Ours, particle-based & HIWYN, $N\!=\!8$ & Bilinear & Bicubic & Nearest Neighbor \\
    \hline
    Moran's $I$ & 5.103e-4 / 0.849 & -1.995e-3 / 0.475 & 3.215e-3 / 0.243 & 0.612 / 0 & 0.983 / 0 & 2.974e-2 / 6.103e-27\\
    \hline
    K-S Test & 3.410e-3 / 0.430 & 3.023e-3 / 0.586 & 3.274e-3 / 0.482 & 0.366 / 0 & 0.422 / 0 & 9.806e-3 / 6.681e-06
\\ \hline
\end{tabular}
}
    \caption{
    Preservation of Gaussian white noise achieved by different warping methods. We report scores and p-values for both Moran's $I$ (spatial correlation) and K-S test (normality). We show that results from our method (both variants) and HIWYN are indistinguishable from white Gaussian noise, while generic warping methods lead to corrupted noise.}
    \label{fig: white_noise}
\end{figure}

\begin{figure}[!t]
    \raggedright
    \raisebox{5pt}{
    \begin{subfigure}
        \centering
        \includegraphics[width=0.29\linewidth]{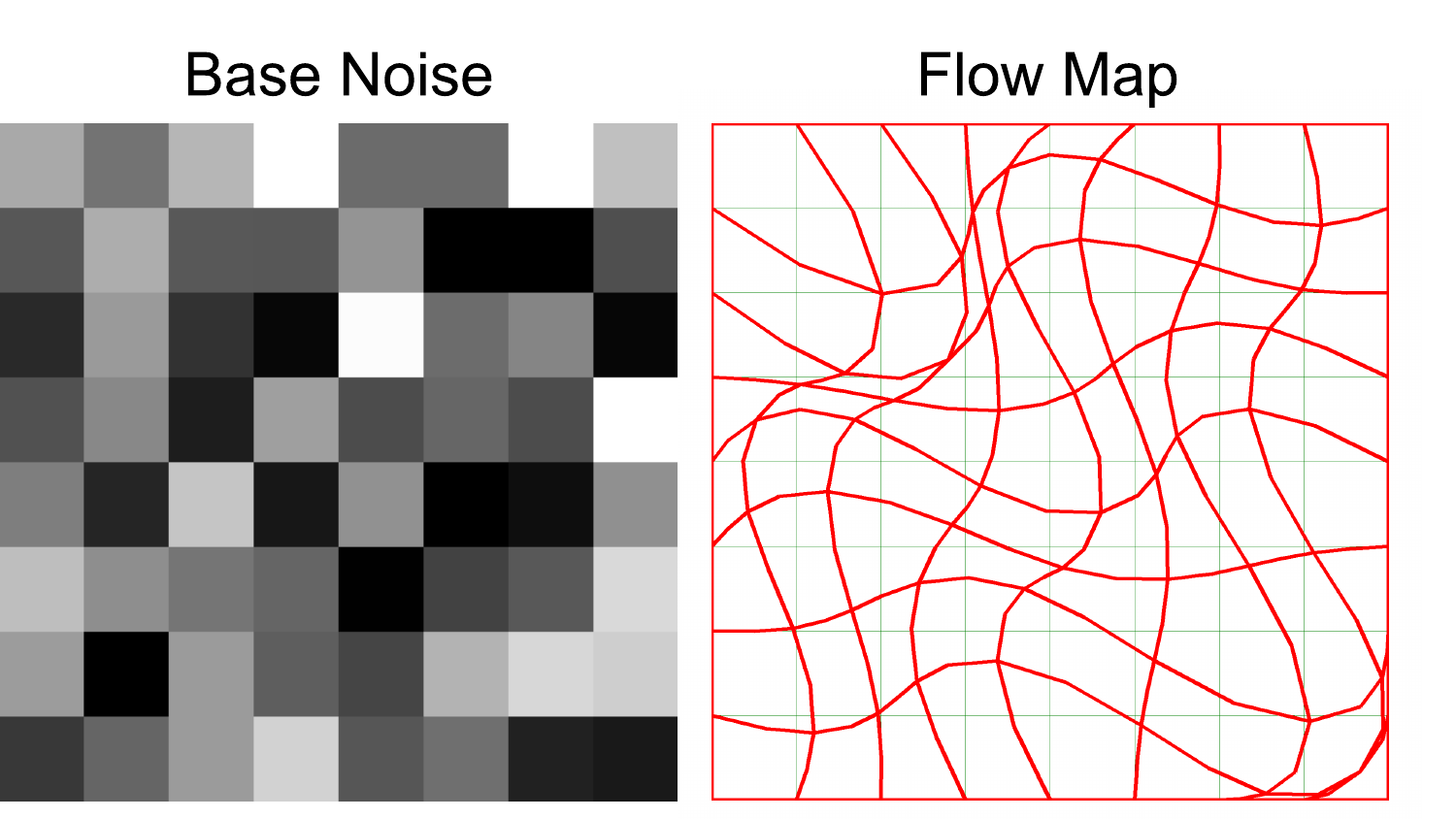}
    \end{subfigure}
    }
    \hspace{-14pt}
    \begin{subfigure}
        \centering
        \includegraphics[width=0.23\linewidth]{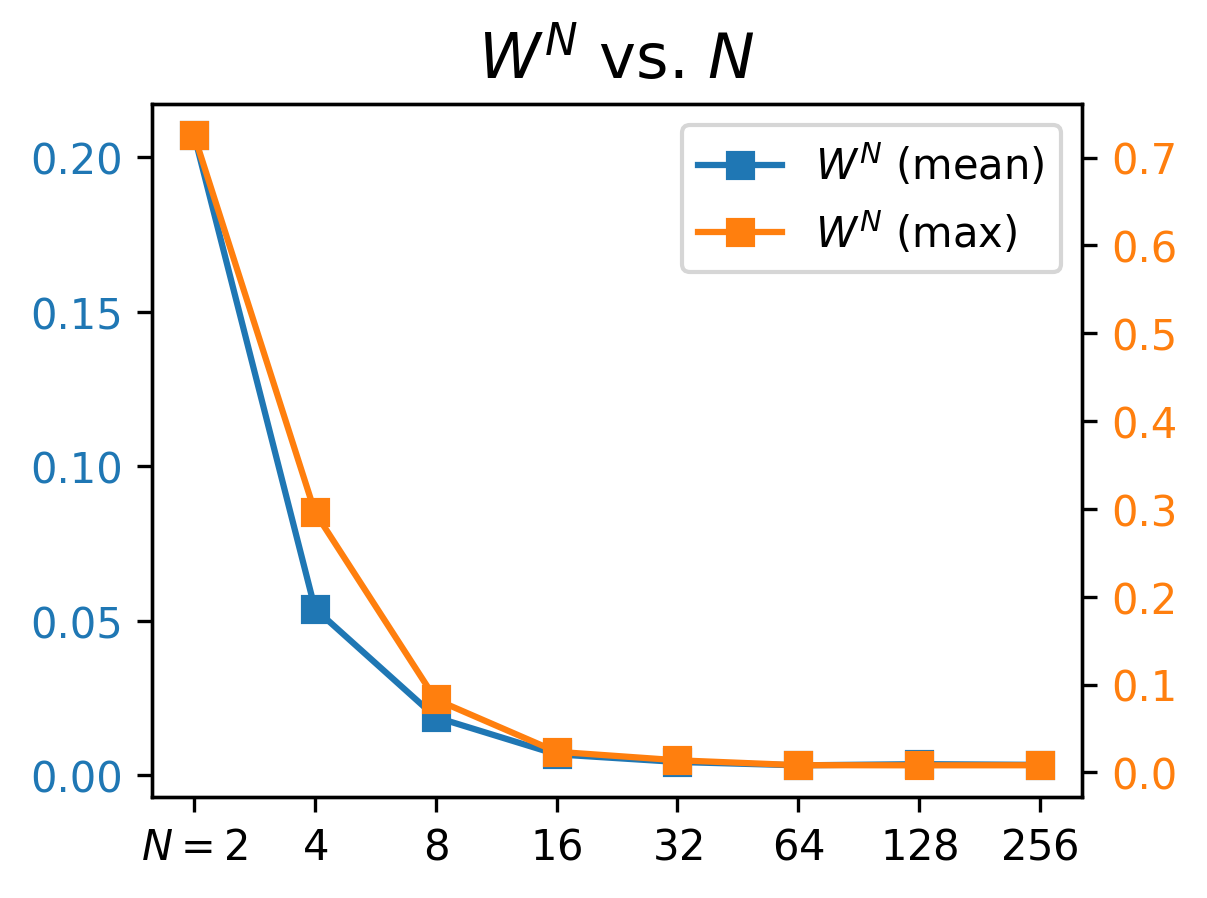}
    \end{subfigure}
    \hspace{-10pt}
    \begin{subfigure}
    \centering
    \raisebox{32pt}{
    \scalebox{0.69}{
    \renewcommand{\arraystretch}{1.2}
    \begin{tabular}{c|c|c||c|c|c}
    \hline
    \multicolumn{6}{c}{$W^{N}$ for Varying Values of $N$}\\
    \hline
    $N$&mean $W^N$&max $W^N$&$N$&mean $W^N$&max $W^N$ \\
    \hline
    2 & 2.072e-1 & 7.253e-1 & 32 & 4.320e-3 & 1.391e-2\\
    \hline
    4 & 5.394e-2 & 2.962e-1 & 64 & 3.236e-3 & 8.325e-3\\
    \hline
    8 & 1.881e-2 & 8.310e-2 & 128 & 3.616e-3 & 8.134e-3\\
    \hline
    16 & 6.792e-3 & 2.361e-2 & 256 & 3.387e-3 & 8.228e-3\\
    \hline
    \end{tabular}
    }
    }
    \end{subfigure}
    \begin{subfigure}
        \centering
        \vspace{-17pt}
        \includegraphics[width=\linewidth]{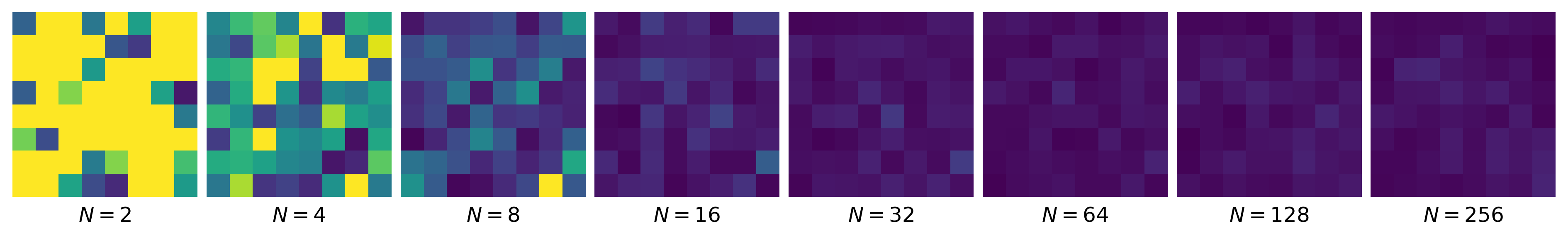}
    \end{subfigure}
    \vspace{-0.25in}
    \caption{Convergence of HIWYN to our method as $N$ increases. Top left: experimental setup with prior noise and deformation map. Top middle: 2-Wasserstein distance $W^N$ between the output of HIWYN and ours. Top right: statistics table. Bottom: $W^N$ difference image between the output of HIWYN and ours as $N$ increases.
    Notice $W^N$ becomes statistically insignificant for $N \ge 64$. 
    }
    \label{fig:convergence}
\end{figure}

\textbf{Gaussian White Noise Preservation.}
In \Cref{fig: white_noise},
we iteratively warp a noise image
by the same deformation map for 50 timesteps. 
We gauge the output noise's resemblance to Gaussian white noise by measuring normality using one-sample Kolmogorov-Smirnov (K-S) test and detecting spatial correlation using Moran's $I$.
Our results show that %
both HIWYN and ours generate noise images indistinguishable from Gaussian white noise while baseline warping methods cannot pass either test.

\textbf{Convergence of \cite{chang2024warped}.}
We validate that our method (grid-based) is the limiting case of HIWYN. 
Starting with an $8 \times 8$ prior noise image and a flow map (\Cref{fig:convergence}, top left), we run our method and HIWYN with upsampling resolutions $N \in \{2, 4, 8, \ldots, 256\}$ for \num{100000} independent runs to estimate the warped noise's distribution.
For each $N$, we compute the 
2-Wasserstein distance $W^N$ between the output distribution of HIWYN and that of our method. 
The results in \Cref{fig:convergence} demonstrate the convergence of HIWYN to our method as $N$ increase, and reveal that $N\!=\!8$ (recommended by \cite{chang2024warped}) is not yet in the converged phase %
to yield a negligible $W^N$. %

\begin{figure}[!t]
    \vspace{-15pt}
    \begin{subfigure}
    \centering
    \raisebox{48pt}{
    \renewcommand{\arraystretch}{1.2}
\scalebox{0.82}{
\begin{tabular}{c|c|c|c}
\hline
\multicolumn{4}{c}{Comparison of Time and Memory Costs}\\
\hline
    Method     & Time (CPU) & Time (GPU) & Memory \\
\hline
$N = 2$ (HIWYN)         & 19.30s & 2.597s & 293.7MB \\         
\hline
$N = 4$ (HIWYN)         & 75.33s & 9.247s & 746.6MB \\
\hline
$N = 8 $ (HIWYN)         & 398.3s & 35.91s & 2147MB \\
\hline
$N = \infty$ (ours, grid)         & 20.16s & 4.491s & 232.9MB \\
\hline
$N = \infty$ (ours, particle)         &  15.68s & 0.862s & 320.9MB
\\ \hline
\end{tabular}
}
    }
    \end{subfigure}
    \begin{subfigure}
        \centering
        \includegraphics[width=0.34\linewidth]{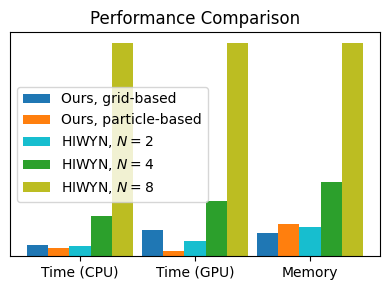}
    \end{subfigure}
    \vspace{-0.2in}
    \caption{Runtime and memory usage of our method vs. HIWYN with $N\!=\!2, 4, 8$. We compare total allocated memory and kernel time on a CPU/GPU. The computation is done on a laptop with Intel i7-12700H and Nvidia RTX 3070 Ti. } 
    \label{fig: performance}
\end{figure}

\textbf{Performance Comparison.}
For our method (both variants) and HIWYN with upsampling levels $N \in \{2, 4, 8\}$, we perform 100 independent runs on a $1024\times 1024$ image. 
We report the kernel time 
with CPU and GPU backends (\Cref{fig: performance}) as well as the memory usage. %
The runtime and memory usage of both our variants %
are largely comparable to those of HIWYN with $N\!=\!2$.
Compared to HIWYN with $N\!=\!8$, both our variants offer order-of-magnitude improvements in runtime and memory usage. 
Specifically, our grid-based variant extends HIWYN to infinite upsampling resolution while being $19.7\times$ faster on CPU and $8.0\times$ faster on GPU, using $9.22\times$ less memory; and our particle-based method, %
albeit not strictly equivalent to HIWYN at $N\!=\!\infty$, 
achieves a $41.7\times$ speedup on GPU. 
In the following sections, we show that our particle-based variant consistently achieves comparable quality to the grid-based variant in real-world scenarios (see video results). 

\begin{figure}[!t]
    \begin{subfigure}
        \centering
        \includegraphics[ width=0.98\linewidth]{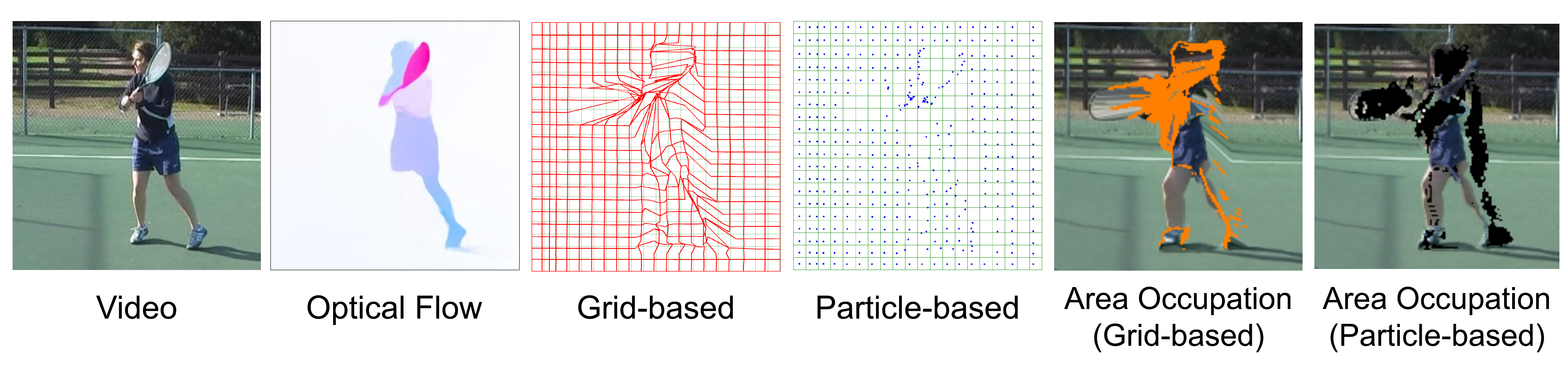}
    \end{subfigure}
    \vspace{-0.1in}
    \caption{Comparison of grid-based vs. particle-based variants under non-diffeomorphic optical flow. Pixels with detected overlaps are colored in orange. Further results are given in Figure~\ref{fig:mesh_free_nondiffeomorphic}.}
    \label{fig: nondiffeomorphic_simplified}
\end{figure}

\begin{figure}[!t]
\vspace{-10pt}
    \begin{subfigure}
        \centering
        \includegraphics[ width=0.98\linewidth]{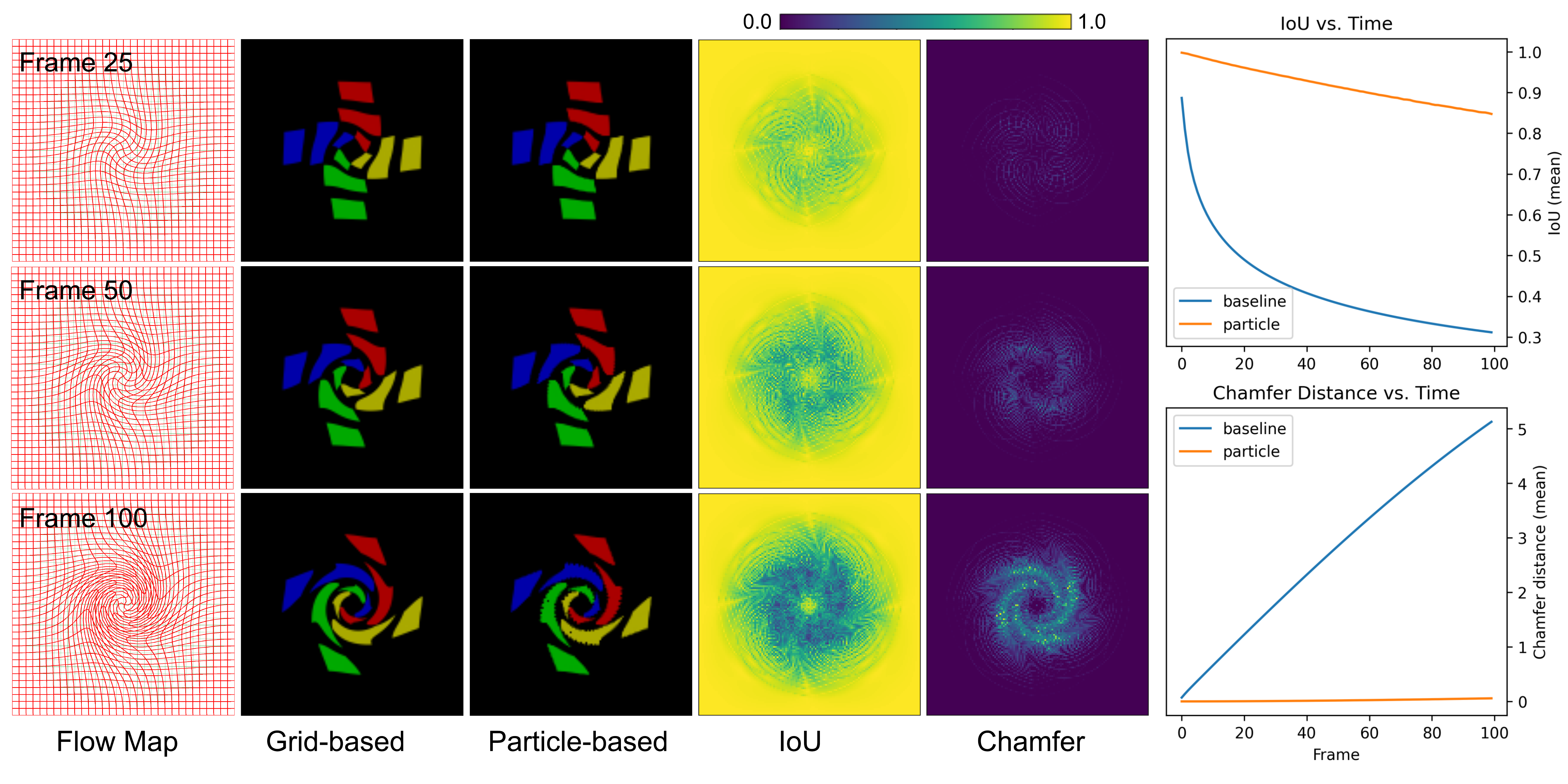}
    \end{subfigure}
    \vspace{-0.1in}
    \caption{Comparison between the grid-based and particle-based variants for building partition records when the deformation map is diffeomorphic. The first column shows the deformation map at different frames. The second and third columns visualize warped pixel regions by the two variants. 
    The fourth and fifth columns show IoU (larger is better) and Chamfer distance (smaller is better) between these deformed pixel regions.
    In the two plots on the right, we contextualize and visualize how much the two variants differ by plotting their IoU and Chamfer distance against those computed between the grid-based variant and the identity map, which shows that the particle-based variant remains close to the grid-based variant even under large distortion.}
    \label{fig: mesh_free_diffeomorphic}
    \vspace{-7pt}
\end{figure}

\textbf{Comparison between Grid-Based and Particle-Based Variants.}
In \Cref{fig: mesh_free_diffeomorphic}, we compare both variants when the deformation map %
is diffeomorphic under different levels of distortion. 
Visually, the difference between the two variants is negligible at frame 25 and becomes noticeable at frame 100. %
We measure this difference by comparing the deformed regions for each pixel in terms of IoU and weighted Chamfer distance.
We additionally compare the particle-based result with that of an identity-map baseline (right column in \Cref{fig: mesh_free_diffeomorphic}), which shows that the gap between the two variants remains small even under large distortion.
In \Cref{fig: nondiffeomorphic_simplified}, we stress test both variants under non-diffeomorphic maps obtained using optical flow \citep{teed2020raft} on a real-world video \citep{Bro11a}. In images 3 and 4, we see that the real-world flow map induces inverted meshes for the grid-based variant and clustered particles for the particle-based variant. %
While clustered particles are guaranteed to be assigned disjoint regions as prescribed by \Cref{alg:mesh_free}, mesh inversions cause noise contention issues due to polygon overlaps. In images 5 and 6, we mark the grid cells with noise contention in orange, which occurs in the grid-based variant but not in the particle-based variant.

\begin{figure}[!t]
\vspace{-20pt}
\centering
    \begin{subfigure}
        \centering
        \includegraphics[ width=0.95\linewidth]{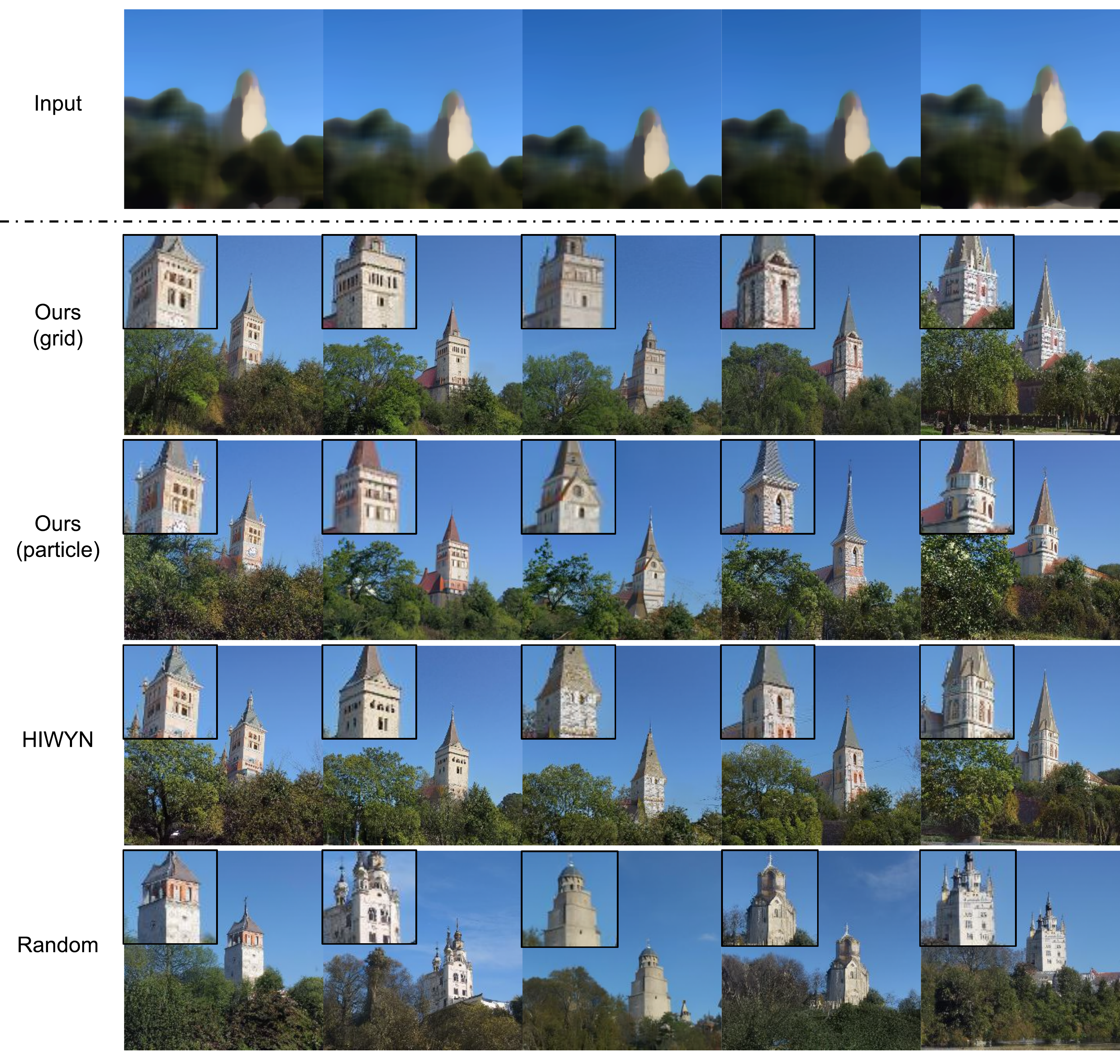}
    \end{subfigure}
    \vspace{-0.1in}
    \caption{We compare the temporal consistency with different noise initialization schemes. We obtain the image sequence using SDEdit with the conditional signal shown at the top. The results of our method (both variants) and HIWYN are shown here, and full results with additional baselines and benchmarks are shown in Figure~\ref{fig: church_all}. We highlight the details of the tower with the inset.}
    \label{fig: church_closeup}
\end{figure}

\textbf{Conditional Video Generation.}
We apply our method to conditional video generation by adapting SDEdit \citep{meng2021sdedit}, a conditional image generation method, to produce temporally consistent video frames. 
We apply Perturbed-Attention Guidance \citep{ahn2024self} to the unconditional models with scale 3.0. %
Our two inputs are a conditioning video (generated by applying a median filter to real-world videos following \cite{chen2023control}) and an optical flow field estimated using RAFT \citep{teed2020raft}. 
Without noise manipulation, if we run SDEdit frame-by-frame (\Cref{fig: church_closeup}, bottom row), the details (\textit{e.g.}, in the tower and trees) would result in strong flickering.
By warping the noise using the optical flow, the temporal consistency is much improved. 
As shown in \Cref{fig: church_closeup}, our methods (both variants) and HIWYN yield comparable visual qualities. 
Full experiments that shows comparison with Control-A-Video \citep{chen2023control} and PYoCo \citep{ge2023preserve} and additional baselines are provided in \Cref{fig: church_all,fig: cat_all} with generation quality metrics reported in \Cref{tab: cat_church_stats}.
We refer to our supplementary video for better visualization.

\textbf{3D Noise Warp.} 
We extend our particle-based algorithm 
to 3D by replacing the bilinear kernel 
with a bicubic kernel in \Cref{alg:mesh_free} %
and apply it to %
GaussianCube \citep{zhang2024gaussiancube}, which denoises a dense 3D noise grid to reconstruct 3D Gaussians. 
We adapt it to perform conditional generation a la SDEdit. Starting with a 3D pickup truck generated unconditionally, we condition the model to generate vehicles with smaller and larger cabins by deforming the truck with a horizontal shear velocity field. We compare the results from using random noise to those using noise warped with our particle-based method. Using the warped noise improves the consistency, reducing the flickering of the cars' geometries and textures. We show the results in \Cref{fig:3D} and refer to our supplementary video for better visualization.

\section{Related Works}

\textbf{Noise in Diffusion Models.}  
Diffusion models generate images from input noise, and noise can thus be considered the counterpart to the latent codes utilized in GAN models. As such, the outputs of diffusion models have dependencies and correlations to the initial input noise, making noise a useful handle to control temporal consistency \citep{khachatryan2023text2video}.
In addition to \cite{chang2024warped} which this work was inspired by and improves upon, there are various other temporal noise manipulation techniques that do not preserve Gaussian noise distribution-- some methods (\cite{ma2024cinemo, ren2024consisti2v}) blend high frequency Gaussian noise with low frequency motion, while others (\cite{mokady2022null, wallace2022edict}) rely on approximating the inversion of noise from temporally coherent image sequences. \cite{pandey2024diffusion} goes one step further and manipulates inverted noise in 3D space. These approaches are flexible but degrade the output of the diffusion model due to the domain gap between inference time noise and training time noise, and as such, have occasionally been accompanied by mitigation strategies such as anisotropic diffusion (\cite{yu2024constructing}).
Noise manipulation is also not limited to the generation and stylization of videos, but has various applications in image editing (\cite{Hou_2024_CVPR, pandey2024diffusion}) and 3D mesh texturing (\cite{richardson2023texture}) as well.

\textbf{Noise in Computer Graphics.} While our noise warping work draws main inspiration from simulation techniques, spatial noise manipulation has been extensively studied in the graphics community through  applications in animation and rendering. Works like \citep{kass2011coherent,burley2024dynamic} present 2D noise manipulation techniques that add a stylized organic hand-drawn look to computer-generated animation via dynamic noise textures \citep{perlin1985image}. In order to make sure the stylization is temporally consistent and visually pleasing, noise textures are deformed in a way that makes them consistent with the underlying animation, but little emphasis is given to the preservation/rigor of the noise distribution. On the other hand, properties of 2D spatial noise have been extensively and rigorously studied in rasterization and raytracing literature \citep{cook1986stochastic,lagae2008comparison}, originating from the idea of using dithering to reduce banding and quantization artefacts in image signal processing \citep{roberts1962dither}. In particular, the lack of low frequency details and clumping in blue noise as opposed to white Gaussian noise has made it the choice of foundational antialiasing methods such as Poisson disc sampling \citep{mccool1992hierarchical}, and recent progress made in this line of antialiasing research has close ties with our methodology. For example, 
\cite{wolfe2022blue} look at accelerating rendering tasks by extending spatial blue noise to the temporal domain, while \cite{huang2024blue} show promising results in supplementing white noise with blue noise during diffusion model training.

\section{Conclusions}

In this paper, we presented \textit{infinite-resolution integral noise warping}, a novel algorithm for computing temporally coherent, distribution-preserving noise transport to guide diffusion models into generating consistent results. By deriving a continuous-space analogy to the discrete, upsampling-based strategy of the current state-of-the-art \citep{chang2024warped}, our method not only further improves the accuracy by effectively raising the upsampling resolution to infinity, but also drastically reduces the computational cost, processing high-resolution noise images in real-time, which removes its main limitation. We also highlight the connotations of our new perspective beyond the performance gains, as it facilitates agnosticism to non-injective maps and extensibility to higher dimensions.

Our work may be extended in a few directions. First, our particle-based variant does not capture temporal correlations induced by contraction or expansion, which may be addressed in the future with Voronoi partitioning.
Secondly, although we only show use cases that leverage flow maps for temporal consistency, our method can operate on other map types such as UV maps for 3D consistency, which might be explored in future works. Thirdly, the connection between the consistency of the initial noise and that of the generated results remains empirical and invites theoretical justifications. Finally, the efficacy of noise warping for latent diffusion models remains to be investigated.

\section*{Acknowledgement}
We thank Lukas Lepicovsky, Ioan Boieriu, David Michielsen, Mohsen Mousavi, and Perry Kain from Eyeline Studios, for providing data that kickstarted this project and for assisting in shaping our research with practical future use cases. We also thank Austin Slakey for sharing his insights on training diffusion models, and Tianyi Xie for filming his cat for our testing.

\bibliography{iclr2025_conference}
\bibliographystyle{iclr2025_conference}

\appendix
\counterwithin{figure}{section}
\newpage
\section{Proof of Theorem~\ref{theo: convergence}}
\label{append: proof1}
\begin{proof}
By unrolling the definitions, for $t \in [0, 1]$, we have
\begin{align*}
    H_n(t) = S_n^*(t) - tS_n^*(1) + tc, \qquad S_n^*(t) \coloneq \frac{1}{\sqrt{n}}\left(\sum_{i=1}^{\floor{nt}} Z_i + (nt-\floor{nt}) Z_{\floor{nt} + 1} \right).
\end{align*}
By \citet[Theorem 5.22]{morters2010brownian}, $\{S_n^*\}_{n\in\mathbb{Z}_{\ge 1}}$ converges in distribution to $W(t)$ under the sup-norm metric of $C[0,1]$.
To lift this convergence to the sequence $\{H_n\}_{n\in\mathbb{Z}_{\ge 1}}$, observe that the function $g: C[0,1] \to C[0, 1]$ defined by
\begin{align*}
g(x(t)) \coloneq x(t) - t x(1) + tc
\end{align*}
is continuous under the sup-norm metric.
To verify this, suppose $\lim_{n\to\infty}f_n = f$ for $\{f_n\}_{n\in\mathbb{Z}_{\ge 1}}, f \in C[0, 1]$. Then
\begin{align*}
    \|g(f_n) - g(f)\|_\infty &= \sup_{t\in[0, 1]} \left| (f_n(t) - t f_n(1) + tc) - (f(t) - t f(1) + tc) \right| \\
    &\le \|f_n - f\|_\infty + \| f_n(1) - f(1)\| \le 2\|f_n - f\|_\infty \to 0.
\end{align*}
Hence, by the continuous mapping theorem,
\begin{align*}
    g(S_n^*) = H_n \stackrel{d}{\longrightarrow} B(t) - t B(1) + tc.
\end{align*}
To show
\begin{align*}
W(t) - tW(1) + tc = (W(t) \mid W(1) = c),
\end{align*}
first of all, the conditioning $(W(t) \mid W(1) = c)$ is interpreted as the limit of $(W(t) \mid \left|W(1) - c\right| < \epsilon)$ as $\epsilon \to 0$.
Denote $Y(t) \coloneq W(t) - t W(1)$, so that $W(t) = Y(t) + t W(1)$.
Since $\Cov(Y(t), tW(1)) = \Cov(W(t)-tW(1), t W(1)) = t\Cov(W(t), W(1)) - t^2 \Var(W(1), W(1)) = 0$ and that $Y(t), tW(1)$ are jointly Gaussian, they are independent. Therefore,
\begin{align*}
    \lim_{\epsilon \to 0}(W(t) \mid \left|W(1) - c\right| < \epsilon) &= \lim_{\epsilon \to 0}(Y(t) + t W(1)\mid \left|W(1)-c\right| < \epsilon)  \\
    &= Y(t) + \lim_{\epsilon \to 0} (tW(1) \mid \left|W(1)-c\right| < \epsilon) \\
    &= W(t) - t W(1) + tc.
\end{align*}
\end{proof}

\newpage
\section{Additional Results}
\begin{figure}[h]
        \centering
        \includegraphics[ width=0.98\linewidth]{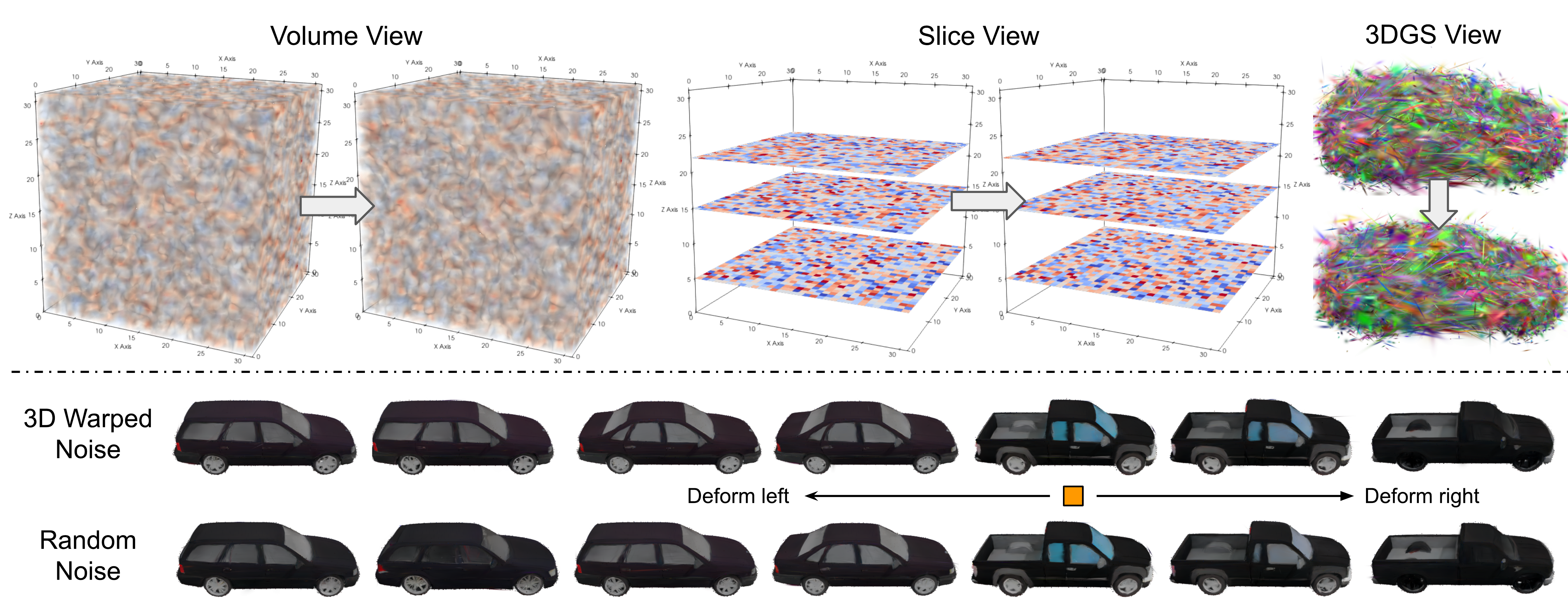}
        
    \caption{Extension of our particle-based variant to perform volumetric noise warping. We show the volume render on the top left, slice views on the top middle, and 3D Gaussians as used in GaussianCube~\citep{zhang2024gaussiancube} on the top right. We show that warping the volumetric noise noticeably facilitates temporal consistency over random baseline when we perform 3D editing, which can be observed from the flickering of the window color on the bottom row when random noise is used. We refer to our supplementary video for better visualization of this result.}
    \label{fig:3D}
\end{figure}

\begin{figure}[t]
\vspace{-0.5in}
\centering
    \begin{subfigure}
        \centering
        \includegraphics[ width=0.95\linewidth]{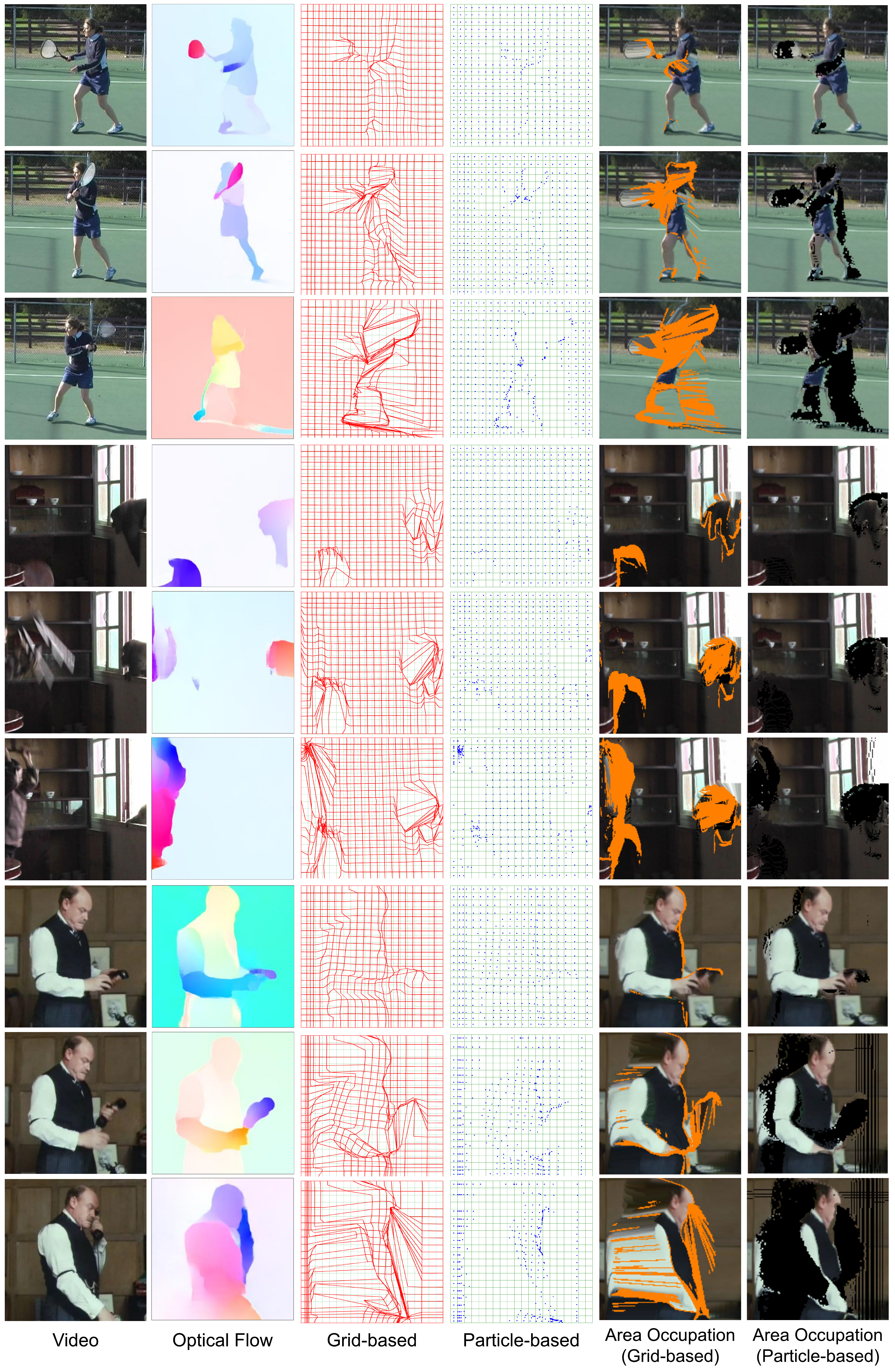}
    \end{subfigure}
    \caption{Comparison of grid-based and particle-based variants under non-diffeomorphic deformation maps generated by optical flow~\citep{teed2020raft}. The orange pixels are the invalid pixels where noise contention occurs. Flow maps are downsampled $10\times$ for better visualization. The test image sequences are borrowed from the dataset by \cite{Bro11a}.}
    \label{fig:mesh_free_nondiffeomorphic}
\end{figure}

\begin{figure}[!t]
\centering
    \vspace{-1.6in}
    \begin{subfigure}
        \centering
        \includegraphics[ width=0.95\linewidth]{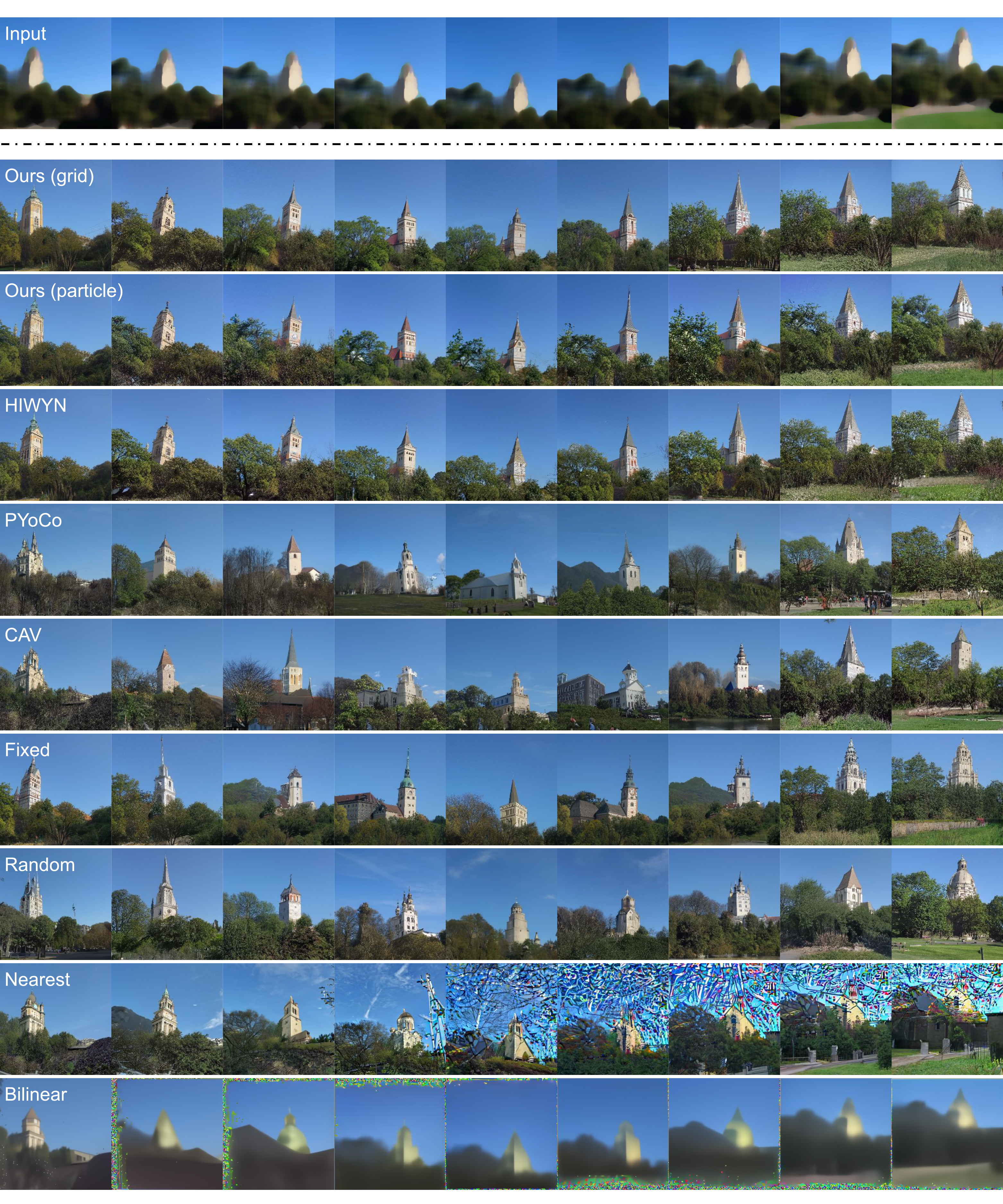}
    \end{subfigure}
    \caption{Results generated by all compared methods on the church scene. On the image level, all methods offer similar generation quality except for the interpolation baselines which yield significantly corrupted results. The difference lies in how the details are preserved across frames. Apart from the details of the main tower, the tree on the bottom left also exposes the interesting differences between noise initialization schemes. We refer to our supplementary video for better visualization of these results.}
    \label{fig: church_all}
\end{figure}

\begin{figure}[t]
\vspace{-1.8in}
\centering
    \begin{subfigure}
        \centering
        \includegraphics[ width=0.95\linewidth]{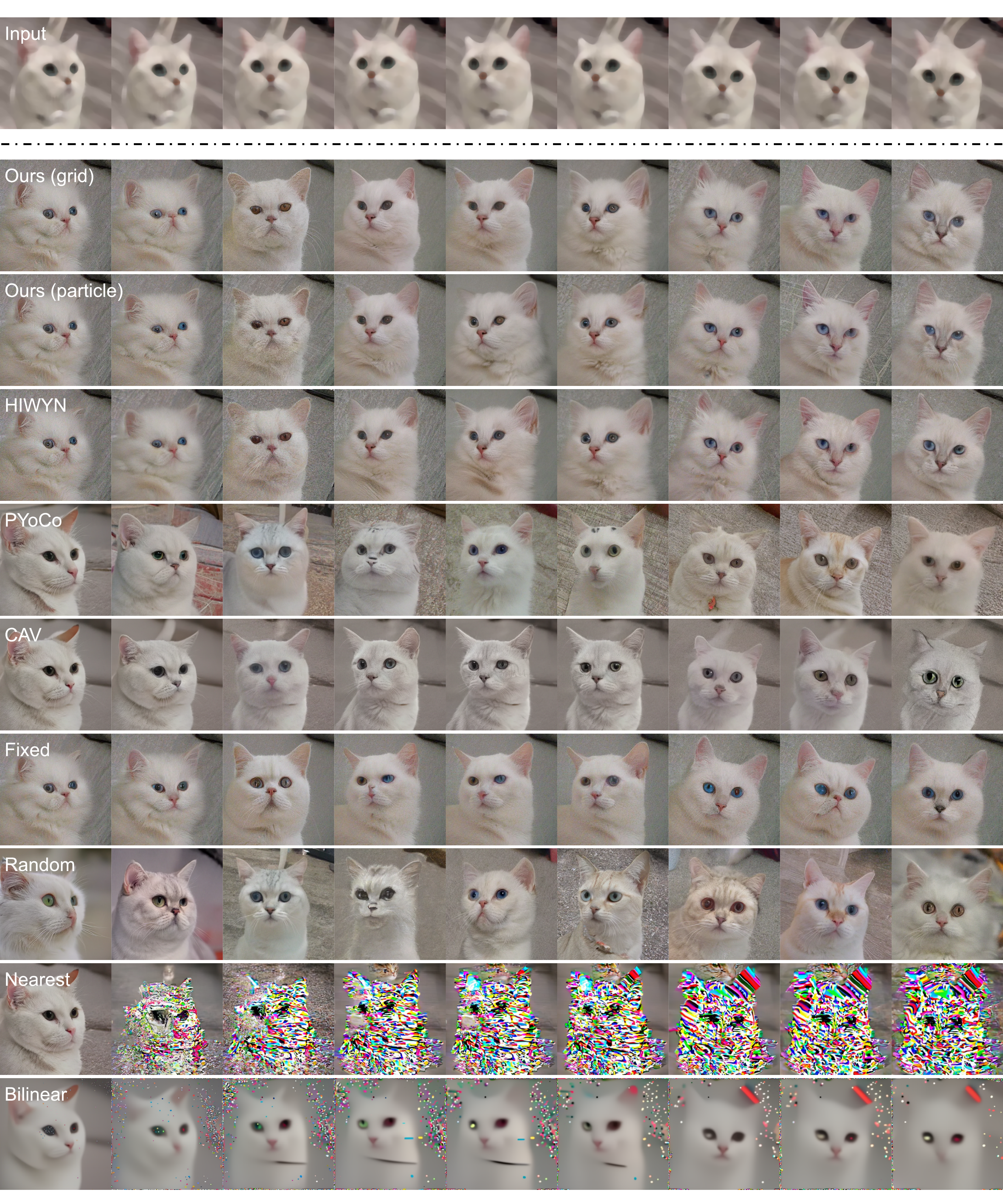}
    \end{subfigure}
    \caption{Results generated by all compared methods on the cat scene. We observe that HIWYN and our method (both variants) yield similar results, which illustrates the appeal of our particle variant due to its efficiency and simplicity. 
    We refer to our supplementary video for better visualization of these results.}
    \label{fig: cat_all}
\end{figure}

In this section, we include additional visual and numerical results. In Figure~\ref{fig: church_all} and Figure~\ref{fig: cat_all}, we showcase extended comparisons results with the addition of Control-A-Video \citep{chen2023control} and PYoCo \citep{ge2023preserve}, along with additional baselines with fixed and interpolated noise using bilinear and nearest interpolating schemes. The corresponding quantitative metrics for both the church and cat scenes are reported in~\Cref{tab: cat_church_stats}. In Figure~\ref{fig:mesh_free_nondiffeomorphic}, we use additional examples to showcase the noise contention issue caused by non-injective meshes that applies similarly to our grid-based variant and \cite{chang2024warped}, and highlight the robustness of our particle-based variant. In Figure~\ref{fig:3D}, we show additional results from combining our particle-based, volumetric noise warp with GaussianCube \citep{zhang2024gaussiancube} to facilitate 3D editing.
\begin{table}[h]
    \centering
    \renewcommand{\arraystretch}{1.2}
\scalebox{0.79}{
\begin{tabular}{c|c|c|c|c|c|c|c|c|c}
\hline
\multicolumn{10}{c}{Video Generation Quality (Church)}\\
\hline
    Metric     & Ours (G) & Ours (P) & HIWYN & PYoCo & CaV & Random & Fixed & Bilinear & Nearest \\
    \hline
    \textit{Consistency $\downarrow$}  & 9.868e-2 & 1.065e-1 & 1.060e-1 & 1.175e-1 & 1.359e-1 & 1.538e-1 & 1.120e-1 & 8.114e-2 & 1.305e-1\\
    \hline
    \textit{Realism $\downarrow$}  & 4.643e-2 &  5.180e-2 & 4.959e-2 &  4.119e-2 &  4.069e-2 &  3.731e-2 &  3.911e-2 &  2.301e-1 & 7.012e-2\\
    \hline
    \textit{Faithfulness $\downarrow$}  & 3.872e-2 & 4.309e-2 & 4.377e-2 & 3.764e-2 & 4.169e-2 & 3.976e-2 & 3.264e-2 & 5.623e-2 & 9.321e-2\\
    \hline
\multicolumn{10}{c}{Video Generation Quality (Cat)}\\
\hline
    Metric     & Ours (G) & Ours (P) & HIWYN & PYoCo & CaV & Random & Fixed & Bilinear & Nearest \\
    \hline
    \textit{Consistency $\downarrow$} & 6.001e-2 & 5.898e-2 & 5.807e-2 & 6.383e-2 & 4.280e-2 & 1.219e-1 & 3.950e-2 & 3.503e-2 & 1.058e-1\\
    \hline
    \textit{Realism $\downarrow$} & 1.559e-1 & 1.496e-1 & 1.528e-1 & 1.506e-1 & 1.486e-1 & 1.221e-1 & 1.588e-1 & 3.687e-1 & 3.343e-1 \\
    \hline
    \textit{Faithfulness $\downarrow$}  & 2.039e-2 & 2.064e-2 & 2.022e-2 & 2.023e-2 & 1.817e-2 & 2.077e-2 & 1.972e-2 & 3.809e-2 & 2.201e-1\\
    \hline
    
\end{tabular}
}
    \caption{We show the quality metrics for conditional video generating using SDEdit. The consistency is measured using warp MSE following \cite{chang2024warped}, and the realism and faithfulness are measured following \cite{meng2021sdedit}.}
    \label{tab: cat_church_stats}
\end{table}

\newpage

\newpage

\newpage

\end{document}